\documentclass[runningheads]{llncs}

\usepackage{graphicx}
\usepackage{epsfig}
\usepackage{multirow}
\usepackage{amsmath}
\usepackage{amssymb}
\usepackage{booktabs}
\usepackage{dsfont}
\usepackage{cite}
\usepackage{float}
\usepackage{color}
\usepackage{xcolor}
\usepackage{soul}
\usepackage{float}
\usepackage{url}
\usepackage{rotate}
\usepackage{tikz}
\usepackage[colorlinks=true, citecolor=blue]{hyperref}%
\usepackage{makecell}
\usepackage{multirow}
\usepackage{tabularx}
\usepackage{subfigure}
\usepackage{bm}
\usepackage{comment}
\usepackage{wrapfig}

\usepackage[accsupp]{axessibility}  

\begin{document}
\pagestyle{headings}
\mainmatter
\def\ECCVSubNumber{2633}  

\linespread{0.93}

\title{Uncertainty-aware Multi-modal Learning via Cross-modal Random Network Prediction}

\titlerunning{Cross-modal Random Network Prediction}
%
\author{Hu Wang\inst{1} \and
Jianpeng Zhang\inst{2} \and
Yuanhong Chen\inst{1} \and
Congbo Ma\inst{1} \and
Jodie Avery\inst{1} \and
Louise Hull\inst{1} \and
Gustavo Carneiro\inst{1}}
\authorrunning{H. Wang et al.}
%
\institute{The University of Adelaide, Australia \and
Northwestern Polytechnical University, China
}
\maketitle

\begin{abstract}
Multi-modal learning focuses on training models by equally combining multiple input data modalities during the prediction process. However, this equal combination can be detrimental to the prediction accuracy because different modalities are usually accompanied by varying levels of uncertainty. Using such uncertainty to combine modalities has been studied by a couple of approaches, but with limited success because these approaches are either designed to deal with specific classification or segmentation problems and cannot be easily translated into other tasks, or suffer from numerical instabilities. In this paper, we propose a new Uncertainty-aware Multi-modal Learner that estimates uncertainty by measuring feature density via Cross-modal Random Network Prediction (CRNP). CRNP is designed to require little adaptation to translate between different prediction tasks, while having a stable training process. From a technical point of view, CRNP is the first approach to explore random network prediction to estimate uncertainty and to combine multi-modal data. Experiments on two 3D multi-modal medical image segmentation tasks and three 2D multi-modal computer vision classification tasks show the effectiveness, adaptability and robustness of CRNP. Also, we provide an extensive discussion on different fusion functions and visualization to validate the proposed model\footnote{This project received grant funding from the Australian Government through the Medical Research Future Fund - Public Health Research Development Infrastructure PHRDI 000014 Grant and the Australian Research Council through grants DP180103232 and FT190100525.}.

\end{abstract}

\section{Introduction}

Multi-modal data analysis, where the input data comes from a wide range of sources, is a relatively common task. 
For instance, automatic driving vehicles may take actions based on the fusion of the information provided by multiple sensors. 
In the medical domain, automated diagnosis often relies on data from multiple complementary modalities.
Recently, we have seen the development of successful multi-modal techniques, such as vision-and-sound classification~\cite{chen2021distilling}, sound source localization~\cite{chen2021localizing}, vision-and-language navigation~\cite{wang2020soft} or organ segmentation from multiple medical imaging modalities~\cite{dou2020unpaired, wang2020modality,wang2021transbts}. However, current multi-modal models typically rely on complex structures that neglect the uncertainty present in each modality. 
Although they can obtain promising results under specific scenarios, they are fragile when facing situations where modalities contain high uncertainties due to noise in the data or the presence of abnormal information. 
Such issue can reduce their prediction accuracy and limit their applicability in safety-critical applications \cite{han2021trusted}.

Uncertainty is a crucial issue in many machine learning tasks because of the inherent randomness of machine learning processes.
For instance, the randomness of data collection, data labeling, model initialization and training are sources of uncertainty that can result in large disagreements between models trained under similar conditions.
According to~\cite{gawlikowski2021survey, abdar2021review, kendall2017uncertainties}, total uncertainty comprise: 
1) aleatoric uncertainty (also known as data uncertainty), representing inherent noise in the data due to issues in data acquisition or labeling; and 
2) epistemic uncertainty (i.e., model or knowledge uncertainty), which is related to the model estimation of the input data that may be inaccurate due to insufficient training steps/data, poor convergence, etc. 
Total uncertainty is defined as:
\begin{equation} \label{uncertainty} \small
    \underbrace{\mathbb{D}_{p(y \mid x, \theta)}[y]}_{\text {Total Uncertainty}} = \ \underbrace{\mathbb{E}_{p(\theta | D)}\left[\mathbb{D}_{p(y \mid x, \theta)}[y]\right]}_{\text {Aleatoric Uncertainty }}+\underbrace{\mathbb{D}_{p(\theta | D)}\left[\mathbb{E}_{p(y \mid x, \theta)}[y]\right]}_{\text {Epistemic Uncertainty }},
\end{equation}
where $D$ indicates the given dataset, $x$ and $y$ are the inputs and outputs of the model, and $\mathbb{D}[\cdot]$ represents the measurement of disagreement (e.g., entropy). 
The estimation of aleatoric uncertainty is considered as the expectation of the predicted disagreement for each model on data points posterior parameterized by $\theta$; while the epistemic uncertainty is shown by the disagreement of different models parameterized by $\theta$ sampled from the posterior. In this paper, we focus on estimating total uncertainty.

In multi-modal methods, existing methods typically assume that each modality contributes equally to the prediction outcome~\cite{nie2016fully,valindria2018multi,dou2020unpaired}. 
This strong assumption may not hold 
if one of the modalities leads to a highly uncertain prediction, which can damage the model performance. 
In general, deep learning models that can estimate uncertainty~\cite{kohl2018probabilistic,kohl2019hierarchical,baumgartner2019phiseg} were not designed to deal with multi-modal data. 
These models are usually based on Bayesian learning that have slow inference time and poor training convergence, or on abstention mechanisms~\cite{sensoy2018evidential} that may suffer from the low  representational power of characterising all types of  uncertainties with a single abnormal class. 
Recently, there have been a couple of methods designed to model multi-modal uncertainty~\cite{han2021trusted,monteiro2020stochastic}, but they are  limited to work with very specific classification and segmentation problems, or they show numerical instabilities. 

In this paper, we propose a novel approach to estimate the total uncertainty present in multi-modal data by measuring feature density via \underline{C}ross-modal \underline{R}andom \underline{N}etwork \underline{P}rediction (CRNP).
CRNP measures uncertainty for multi-modal Learning using random network predictions (RNP)~\cite{burda2018exploration}, where the model
is designed to be easily adaptable to disparate tasks (e.g., classification and segmentation) and training is based on a stable optimization that mitigates numerical instabilities.
To summarize, the main contributions of this paper are:
\begin{itemize}
\item We propose a new uncertainty-aware multi-modal learning model through a feature distribution learner based on RNP, named as Cross-modal Random Network Prediction (CRNP). 
CRNP is designed to be easily adapted to disparate tasks (e.g. classification and segmentation) and to be robust to numerical instabilities during optimization.
\item This paper introduces a novel uncertainty estimation based on fitting the output of an RNP, which from a technical viewpoint, represents a departure from more common uncertainty estimation methods based on Bayesian learning or abstention mechanisms.
\end{itemize}

The adaptability of CRNP is shown by its application on two 3D multi-modal medical image segmentation tasks and three multi-modal 2D computer vision classification tasks, where the proposed model achieves state-of-the-art results on all problems.
We perform a thorough analysis of multiple CRNP fusion strategies and present visualization to validate the effectiveness of the proposed model.

\section{Related Work}
\subsection{Multi-modal Learning}

Multi-modal learning has attracted increasing attention from computer vision (CV) and medical image analysis (MIA). 
In MIA, Jia et al.~\cite{jia2020semi} introduced a shared-and-specific feature representation learning for semi-supervised multi-view learning. 
Dou et al.~\cite{dou2020unpaired} proposed a chilopod-shaped multi-modal learning architecture with separate feature normalization for each modality and a knowledge distillation loss function. In CV, Shen et al.~\cite{chen2021localizing} defined a trusted middle-ground for video-and-sound source localization. 
In video-and-sound classification, Chen et al.~\cite{chen2021distilling} proposed to distill multi-modal image and sound knowledge into a video backbone network through compositional contrastive learning. 
Also in video-and-source classification, Patrick et al.~\cite{patrick2020multi, patrick2021space} brought the idea of self-supervision learning into multi-modal by training the networks on external data, which boosted classification accuracy greatly. 
By exchanging channels, Wang et al.~\cite{wang2020deep} showed that the multi-modal features are able to fuse in a better manner. 
Analyzing existing multi-modal learning methods, even though successful on several tasks, they do not consider that when reaching a decision, some modalities may be more reliable than others, which can damage the accuracy of the model.

\subsection{Uncertainty-based Learning Models}

Uncertainty also has been widely studied in deep learning. Corbiere et al.~\cite{corbiere2019addressing} proposed to predict a single uncertainty value by an external confidence network via training on the ground-truth class. 
Sensoy et al.~\cite{sensoy2018evidential} introduced the Dirichlet distribution for an overall classification uncertainty measurement based on evidence. Kohl et al.~\cite{kohl2018probabilistic} proposed a probabilistic UNet segmentation architecture to optimize a variant of the evidence lower bound (ELBO) objective. 
Based on the probabilistic UNet model, Kohl et al.~\cite{kohl2019hierarchical} and Baumgartner et al.~\cite{baumgartner2019phiseg} further updated the model in a hierarchical manner from either the backbone network or prior/posterior networks. 
Jungo et al.~\cite{jungo2019assessing}
used two medical datasets to compare 
several uncertainty measurement models, namely: softmax entropy \cite{gal2016dropout}, Monte Carlo dropout \cite{gal2016dropout}, aleatoric uncertainty \cite{kendall2017uncertainties}, ensemble methods \cite{lakshminarayanan2017simple} and auxiliary network \cite{devries2018leveraging, robinson2018real}. In MIA, multiple uncertainty measurements have been proposed as well~\cite{wang2021tripled, wang2021medical, luo2021efficient, li2021dual}. However, none of the methods above are designed for multi-modal tasks and some of them contain long and complex pipelines that are not easily adaptable to new tasks. Bayesian or ensemble-based methods demand long training and inference times and have slow convergence. Evidential methods have drawbacks too, where the main issue is the representational power of the abstention class. In contrast, our proposed model, by introducing random network fitting for cross-modal uncertainty measurement, is not only technically novel, but it is also simple and easily adaptable to many tasks without requiring any restrictive assumption about uncertainty representation.

\subsection{Combining Uncertainty and Multi-modal Analysis}

Some methods have studied the combination of uncertainty modeling and multi-modal learning. 
For example, 
a trusted multi-view classification model has been developed by modeling multi-view uncertainties through Dirichlet distribution and merging multi-modal features via Dempster's Rule~\cite{han2021trusted}. 
However, it is rigidly designed for classification problems, and cannot be easily translated to other tasks, such as segmentation. Monteiro et al.~\cite{monteiro2020stochastic} took pixel-wise coherence into account by optimizing low-rank covariance metrics to apply on lung nodules and brain tumor segmentation. 
Nevertheless, the method by Monteiro et al.~\cite{monteiro2020stochastic} requires a time-consuming step to generate binary brain masks to remove blank areas, 
and the method is also numerically unstable when training in areas of infinite covariance such as the air outside the segmentation target\footnote{As stated by SSN implementation \cite{monteiro2020stochastic} at  \url{https://github.com/biomedia-mira/stochastic\_segmentation\_networks}.}. 
From an implementation perspective, this method~\cite{monteiro2020stochastic} is also memory intensive when indexing the identity matrix to create one-hot encodings. Differently, in our model, the uncertainty is measured by modeling the overall distribution directly from features without constructing any second-order relation matrix, leading to a numerically more stable optimization and a smaller memory consumption.

\section{Cross-modal Random Network Prediction}

Below, we first introduce the Random Network Prediction (RNP), with a theoretical justification for its use to measure uncertainties. Then we present the CRNP model training and inference with the cross-modal uncertainty measuring mechanism to take the RNP uncertainty prediction from one modality to enhance or suppress the outputs for other modalities when producing a classification or segmentation prediction. 

\subsection{Random Network Prediction}
\vspace{-1mm}

\begin{figure}[t]
\begin{center}
\includegraphics[width=0.7\textwidth]{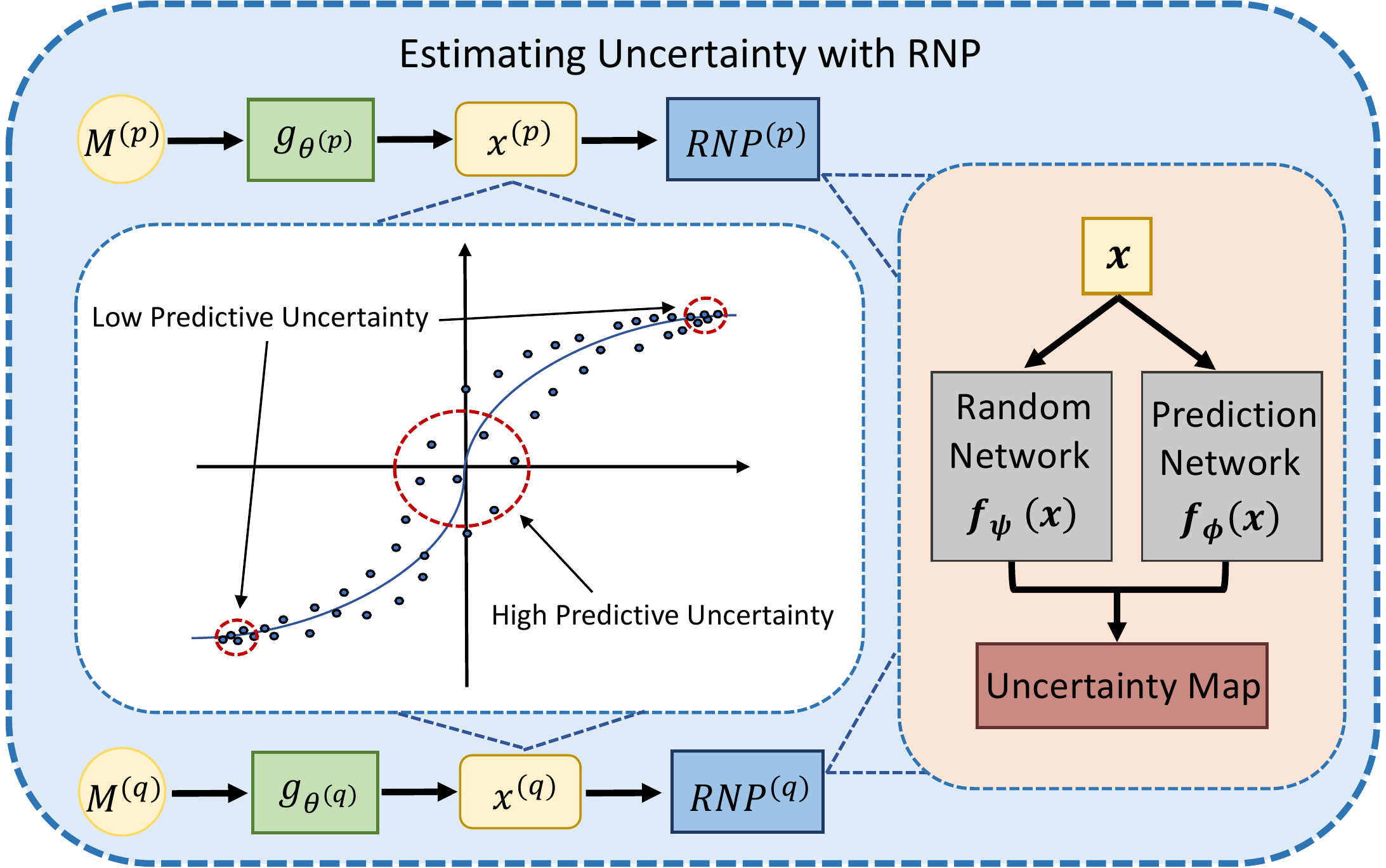}
\end{center}
\vspace{-5mm}
\caption{
The input data $M^{(p)}$ and $M^{(q)}$ are first processed by backbone models $g_{\theta^{(p)}}$ and $g_{\theta^{(q)}}$ that produce the features $x^{(p)}$ and $x^{(q)}$. Then the RNP modules have a fixed-weight random network $f_{\psi}(x)$ and a learnable prediction network $f_{\phi}(x)$ that tries to fit the output of the random network. The prediction network will fit better (i.e., with low predictive uncertainty) at more densely populated regions of the feature space, as shown in the graph. Hence, the difference between the outputs by $f_{\psi}(x)$ and $f_{\phi}(x)$ can be used to estimate uncertainty when processing a test input data.
}\label{fig:crnp-training}
\vspace{-3mm}
\end{figure}

The uncertainty of a particular modality is estimated with the RNP
depicted in Fig.~\ref{fig:crnp-training}.
Specifically, for each RNP, we train a prediction network to fit the outputs of a weight-fixed and randomly-initialized network for feature density modeling. 
The intuition is that the prediction network will fit better the random network outputs of samples (i.e., with low uncertainty), populating denser regions of the feature space; but the fitting will be worse (i.e., with high uncertainty) for samples belonging to sparser regions.
This phenomenon is depicted in the graph inside Fig.~\ref{fig:crnp-training}. 

Formally, 
we consider input images from two modalities $M^{(p)},M^{(q)} \in \mathcal{M}$, where $p$ and $q$ represent the modalities.
After the input image $M^{(p)}$ pass through the encoder $g_{\theta^{(p)}}:\mathcal{M} \to \mathcal{X}$ (similarly for $g_{\theta^{(q)}}(.)$), the features of the two modalities $x^{(p)},x^{(q)} \in \mathcal{X} \subset \mathbb{R}^{N}$ are analyzed by each RNP module. 
The RNP module feeds $x^{(p)}$ and $x^{(q)}$ to a randomly initialized neural network $f_{\psi}:\mathcal{X \to \mathcal{Z}}$, where $\mathcal{Z} \subset \mathbb{R}^M$,
with fixed weights $\psi \in \Psi$. Meanwhile, $x^{(p)}$ and $x^{(q)}$ are fed to a learnable prediction network $f_{\phi}:\mathcal{X} \to \mathcal{Z}$ with parameters $\phi \in \Phi$.
The prediction network has the same output space but a different structure from the random network, where the capacity of  $f_{\phi}$ is smaller than $f_{\psi}$ to prevent potential trivial solutions. 
The cost function used to train the RNP module is based on the mean square error (MSE) between the outputs of the prediction and random networks:
\begin{equation} \label{eqn:rnp_obj} \small
    \phi^{*}=\arg\min_{\phi} \sum_{i=1}^{n} \ell_{MSE}(f_{\phi}(x_i),f_{\psi}(x_i)) + \mathcal{R}(\phi),
\end{equation}
where $n$ denotes the number of training samples,
$\ell_{MSE}(f_{\phi}(x_i),f_{\psi}(x_i)) = 
    \|f_{\phi}(x_i) - f_{\psi}(x_i)\|_2^{2}$, and $\mathcal{R}(\phi) = \| \phi \|_2^2$.
The cost function in~\eqref{eqn:rnp_obj} provides a simple yet powerful supervisory signal to enable the prediction network to learn the uncertainty measuring function.

\subsection{Theoretical Support for Uncertainty Measurement}

The RNP has a strong relation with uncertainty measurement. 
Let us consider a
regression process from a 
set of perturbed data $\tilde{\mathcal{D}}=\left\{\left(x_{i}, \tilde{y}_{i}\right)\right\}_{i=1}^{n}$. 
Considering a Bayesian setting, the objective is to minimize the distance between the ground truth $\tilde{y}_{i}$ and a sum
made up of a generated prior $f_{\psi}(x_i)$ randomly sampled from a Gaussian and an additive posterior term $f_{\phi}(x_i)$ with a regularization $\mathcal{R}(\phi)$. Formally, the optimization is as follows:
\begin{equation} \small
    \phi^{(*)}=\arg\min_{\phi}\sum_{i=1}^{n}\left\|\tilde{y}_{i}-\left[f_{\psi}(x_i)+f_{\phi}(x_i)\right]\right\|_{2}^{2}+\mathcal{R}(\phi),
    \label{eqn:theory_rnp}
\end{equation}
where, according to Lemma 3 in~\cite{osband2018randomized}, the sum $[f_{\psi}(x_i)+f_{\phi}(x_i)]$ is an approximator of the genuine posterior. 
If we fix the target $\tilde{y}_{i}$ with zeros, then the objective to be optimized would be equivalent to minimize the distance between the posterior $f_{\phi}(x_i)$ and the randomly sampled prior $f_{\psi}(x_i)$. 
Thus, each output element within the randomized function or the predict function can be viewed as a member of a set of weight-shared ensemble functions \cite{burda2018exploration}. The predicted error, therefore, can be viewed as an estimate of the variance of the ensemble uncertainty.

\subsection{Training and Inference of CRNP}
\label{sec:cross-rnp}

\begin{figure}[t!]
\begin{center}
\includegraphics[width=0.75\textwidth]{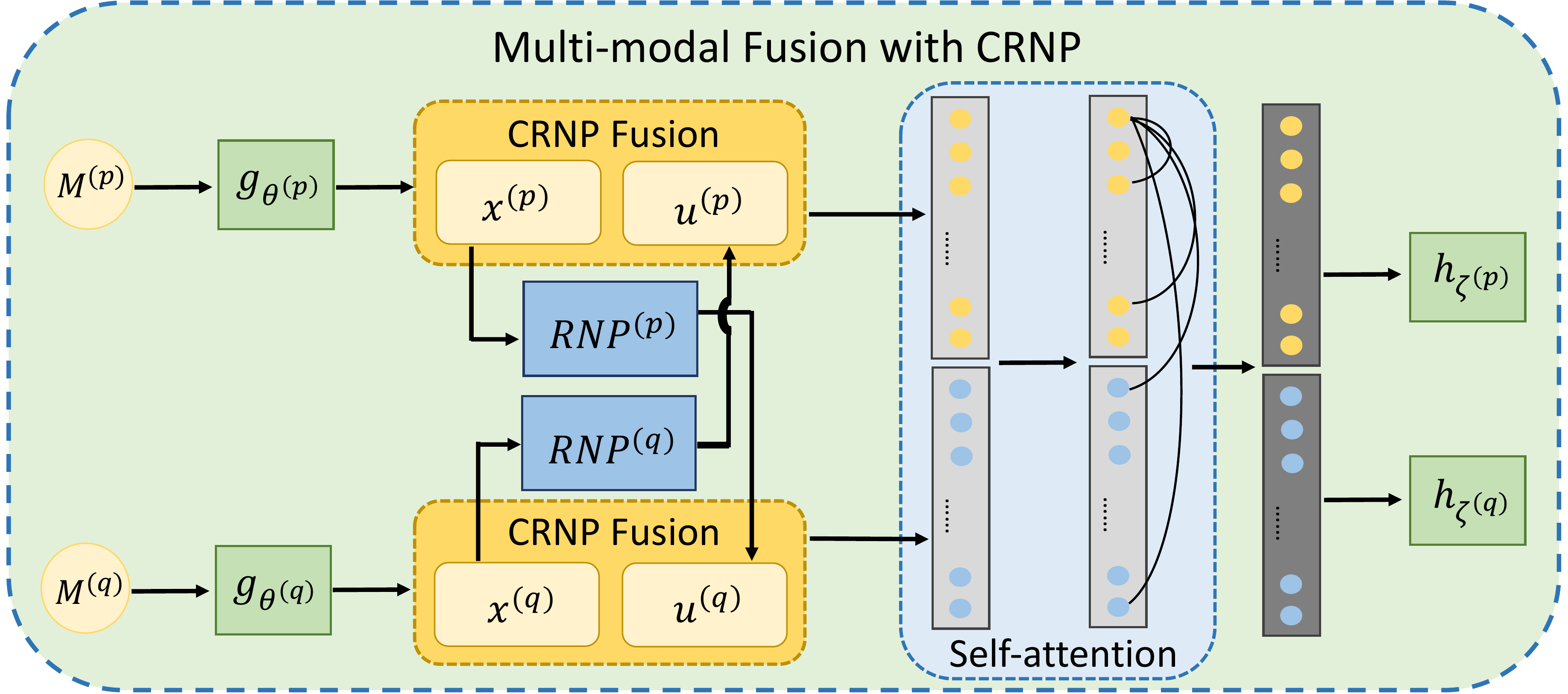}
\end{center}
\vspace{-6mm}
\caption{The overall framework of multi-modal fusion with our CRNP.
}\label{fig:backbone-training}
\vspace{-3mm}
\end{figure}

This section introduces our proposed CRNP, which fuses the multiple modalities with their inferred uncertainties to produce the final predictions (e.g., classification or segmentation), as shown in Fig.~\ref{fig:backbone-training}.
During the multi-modal fusion phase, the features of the two modalities $x^{(p)}$ and $x^{(q)}$ are cross-attended by the uncertainty maps produced by the RNP module from both modalities. The uncertainty map for modality $p$ is represented as:
\begin{equation} \label{eqn:rnp-infer} \small
u^{(p)} = \|f_{\phi^{(q)}}(x^{(q)}) - f_{\psi^{(q)}}(x^{(q)})\|_2^{2},
\end{equation}
and similarly for $u^{(q)}$ for modality $q$. The feature cross-attended by the uncertainty maps is represented by:
\begin{equation} \label{eqn:unet_train} \small
\tilde{x}^{(p)}= \operatorname{fusion}(x^{(p)}, \hat{u}^{(p)} \odot x^{(p)}),
\end{equation}
where $\operatorname{fusion}(.,.)$ represents the operator that fuses the original and cross-attended features, $\hat{u}^{(p)}$ is the channel-wise normalized CRNP uncertainty map, and $\odot$ is the element-wise product operator. 
$\tilde{x}^{(q)}$ is similarly defined as in~\eqref{eqn:unet_train}. Different fusion operations are thoroughly discussed in Sec. \ref{sec:ablation-components}.

We utilize self-attention to further fuse features $\tilde{x}^{(p)}$ and $\tilde{x}^{(q)}$, taking both uni-modal and cross-modal relations between feature elements into consideration. 
As shown in  Fig.~\ref{fig:backbone-training}
, we first concatenate $\tilde{x}^{(p)}$ and $\tilde{x}^{(q)}$ to form the query, key and value inputs for the self-attention module with
$Q=K=V=\operatorname{concatenate}(\tilde{x}^{(p)}, \tilde{x}^{(q)})$.
Then the output of the self-attention is denoted by:
\begin{equation} \small
    l = \operatorname{softmax}\left(\frac{(QW_q) (KW_k^{T})}{\sqrt{d_{k}}}\right) VW_v,
\end{equation}
where $l \in \mathcal{L}$, $W_q$,$W_k$ and $W_v$ are linear projection weights for queries, keys and values, respectively. $d_{k}$ refers to the dimensions of queries, keys and values.
The decoder after the multi-modal fusion is denoted by $h_{\zeta^{(p)}}:\mathcal{L} \to \Delta_{C-1}$ (similarly for $h_{\zeta^{(q)}}$), where $\mathcal{L}$ is the space of the output from the cross-modal RNP module and input to the decoder, and $\Delta_{C-1}$ is the classification simplex (output from softmax).
Note that although the annotations of multi-modal data are similar, they can have significant differences, particularly in segmentation tasks. Hence, without losing generality, we may need to have multiple separate decoders, one for each modality. But multi-decoders are not needed in tasks where the multi-modal annotation is exactly the same. For segmentation problems, the output of $h_{\zeta^{(p)}}$ is the space $\Delta_{C-1}$ per pixel.
The training of CRNP alternates the training of the RNP modules using~\eqref{eqn:rnp_obj} and the training of the whole model.
During RNP training, only the weights of the prediction network inside the RNP are updated by minimising~\eqref{eqn:rnp_obj}, and all other CRNP weights are kept fixed.
During the training of the whole model, all CRNP weights are updated, except for the weights of the prediction network of the RNP.
The whole model training minimizes the multi-class cross-entropy loss for a classification problem or the Dice and element-wise cross-entropy losses for a segmentation model.

During inference, CRNP receives multi-modal inputs, where each modality branch estimates an uncertainty output that will weight the other modality, and the results of both modalities will be fused to produce the final prediction.
CRNP works by assigning large weights to the other modality when the current modality is uncertain.
When both modalities have large uncertainties, the final prediction will rely on a balanced analysis of both modalities.
For the analysis of more than two modalities, the uncertainty map for a particular modality, say $p$, in~\eqref{eqn:rnp-infer} is computed by summing the MSE results produced by all other modalities, with $u^{(p)} = \sum_{q \ne p} \|f_{\phi^{(q)}}(x^{(q)}) - f_{\psi^{(q)}}(x^{(q)})\|_2^{2}$.
The decoders $g_{\theta^{(p)}}(.)$ and $g_{\theta^{(q)}}(.)$ from two modalities can be separated or share-weighted, depending on the corresponding output requirements.

\section{Experiments}
\subsection{Datasets}

\noindent\textbf{Medical Image Segmentation Datasets.} We conduct  experiments on two publicly available multi-modal 3D segmentation datasets: Multi-Modality Whole Heart Segmentation dataset (MMWHS) and Multimodal Brain Tumor Segmentation Challenge 2020 dataset (BraTS2020). 
The MMWHS dataset contains 20 CTs and 20 MRs for training/validation and other 40 CTs and 40 MRs for testing~\cite{zhuang2019evaluation}. Seven classes (background excluded) are considered for each pixel. The two modalities have individual ground-truth (GT) for each CT or MR. 
The BraTS2020 dataset has 369 cases for training/validation and other 125 cases for evaluation, where each case (with four modalities, namely: Flair, T1, T1CE and T2) share one segmentation GT. The evaluation is performed online\footnote{\url{https://ipp.cbica.upenn.edu/categories/brats2020}}. Four classes (background included) are considered for each pixel.

\vspace{1mm}
\noindent\textbf{Computer Vision Classification Datasets.} We also validate our method on three computer vision classification datasets, namely: Handwritten\footnote{\url{https://archive.ics.uci.edu/ml/datasets/Multiple+Features}}, CUB~\cite{wah2011caltech} and Scene15~\cite{fei2005bayesian}. Each sample of the Handwritten dataset contains 2000 samples from six views and it is a ten-class classification problem, 
CUB contains 11,788 bird images from 200 different categories. Following Han et al.~\cite{han2021trusted}, we also adopt the first ten classes 
and two modalities (image and text features) extracted by GoogleNet and doc2vec.
Three modalities are included in Scene15, which contains 4,485 images from 15 indoor and outdoor classes.

\subsection{Implementation Details}

\noindent\textbf{Medical Image Segmentation Tasks.} To keep a fair comparison, the implementation of all models evaluated on MMWHS and BraTS2020 is based on the 3D UNet (with 3D convolution and normalization) as our backbone network. On MMWHS, we adopt the official test set proposed by Zhuang et al.~\cite{zhuang2019evaluation} (40 CTs and 40 MRs) for testing; on BraTS2020, we evaluate all models on the online validation set. For overall performance evaluation, the models were trained for 100,000 iterations on MMWHS and 180,000 iterations on BraTS2020 without model selection. 
Following Dou et al.~\cite{dou2020unpaired}, our hyper-parameter tuning and ablation are conducted on MMWHS with 16 CTs and 16 MRs for training, 4 CTs and 4 MRs for validation. The batch size is set to 2. Stochastic gradient descent optimizer with a momentum of 0.99 is chosen for the model training. 
The initial learning rate is set to $10^{-2}$ on both datasets with cosine annealing \cite{loshchilov2016sgdr} learning rate tuning strategy. 
For the reproduction of Probability UNet~\cite{kohl2018probabilistic}, we use prior/posterior mean instead of random sampling a latent variable $z$ for prediction. 
The evaluation of the methods is based on the Dice score and Jaccard index for MMWHS; and the Dice score and Hausdorff95 index for BraTS2020. 
For cross-modal RNP modules training, the randomized network is made up of 3 depth-wise convolutional hidden layers; the prediction network has 2 depth-wise convolutional hidden layers. 
Between every two layers, both the randomized network and the prediction network adopt Leaky-ReLU as their activation function, where the negative slope is set to $2.5 \times 10^{-1}$. 
We set 256 as RNP output dimension for both tasks. 
For performance evaluation, the CRNP is placed at the bottleneck of our 3D UNet backbone. 
For the ensemble version of CRNP on both datasets, following Wang et al.~\cite{wang2020modality}, we average the logits of 3 CRNP models to reduce the prediction variance.

\vspace{1mm}
\noindent\textbf{Computer Vision Classification Tasks.} For the model evaluation on computer vision datasets, we follow~\cite{han2021trusted} to split the data into 80\% for training and 20\% for testing. To keep a fair comparison, we uniformly trained all models for 500 epochs without model selection and then evaluated them on the test set. The learning rate is set to $3 \times 10^{-4}$; Adam optimizer with $1 \times 10^{-5}$ weight decay and coefficients (0.9, 0.999) are adopted.
Following Han et al.~\cite{han2021trusted}, we apply accuracy and multi-class AUROC as evaluation metrics. We used similar setups for cross-modal RNP modules as on the medical data, with the following differences: the RNP output dimension is set to 32 for computer vision classification tasks and CRNP is placed at the layer before the fully connected layer. 
The training of CRNP model is conducted in an end-to-end manner without any pre-training or post-processing. Also, the hyper-parameters do not require much effort to tune.

\subsection{Medical Image Segmentation Model Performance}
\noindent\textbf{Performance on MMWHS Dataset.}
We compare our approach with: Individual (CT or MR single modality segmentation with separate 3D UNet), 3D UNet (multi-modal fusion by concatenation), the multi-modal learning model Ummkd \cite{dou2020unpaired}, and the uncertainty model Probability UNet\footnote{
We also tried SSN~\cite{monteiro2020stochastic}, but it requires the creation of one-hot encodings that are  memory intensive for seven classes on MMWHS dataset.}~\cite{kohl2018probabilistic}, which proposes a prior net to approximate the posterior distribution, combining the knowledge of inputs and ground truth, in a latent space. 
The evaluation is based on the Dice scores of the segmentation of the left ventricle blood cavity (LV), the myocardium of the left ventricle (Myo), the right ventricle blood cavity (RV), the left atrium blood cavity (LA), the right atrium blood cavity (RA), the ascending aorta (AA), the pulmonary artery (PA) and Whole Heart (WH). All results on MMWHS data 
are obtained by using the official evaluation toolkit\footnote{\url{http://www.sdspeople.fudan.edu.cn/zhuangxiahai/0/mmwhs/}}.

\begin{table}[t!]
\caption{The performance of different models on CT/MR segmentation of MMWHS dataset. The best results for each column within either CT or MR section are in bold. $*$ indicates the result with the ensemble model.}
\vspace{-4mm}
\label{tab:mmwhs}
\begin{center}
\scalebox{0.85}{
\begin{tabular}{l|l|cccccccc} 
\hline
 & Models       & LV              & Myo             & RV              & LA              & RA              & AA              & PA              & WH               \\ 
\hline
                   \multirow{6}{*}{CT} & Individual   & 0.9297          & 0.8943          & 0.8597          & 0.9254          & 0.8701          & 0.9335          & 0.7833          & 0.8989           \\
                    & 3D UNet      & 0.9138          & 0.8781          & 0.8822          & 0.9274          & 0.8680          & 0.9088          & 0.8239          & 0.8957           \\
                    & Ummkd        & 0.9145          & \textbf{0.9066} & 0.8410          & 0.9157          & 0.8853          & 0.8928          & 0.7579          & 0.8734           \\
                    & Prob-UNet    & 0.9071          & 0.8775          & 0.8978          & 0.9262          & 0.8657          & 0.9318          & 0.8425          & 0.8997           \\ 
\cline{2-10}
                    & CRNP (Ours)  & 0.9369          & 0.9036          & 0.9076          & \textbf{0.9375} & 0.8885          & \textbf{0.9538} & 0.8628          & 0.9187           \\
                    & CRNP* (Ours) & \textbf{0.9373} & 0.9060          & \textbf{0.9085} & 0.9366          & \textbf{0.8910} & 0.9503          & \textbf{0.8629} & \textbf{0.9193}  \\ 
\hline \hline
\multirow{6}{*}{MR} & Individual   & 0.8777          & 0.7923          & 0.6146          & 0.5686          & 0.7528          & 0.5854          & 0.3993          & 0.6729           \\
                    & 3D UNet      & 0.8850          & 0.7723          & 0.8559          & 0.8548          & 0.8676          & 0.8551          & 0.7964          & 0.8535           \\
                    & Ummkd        & 0.8721          & \textbf{0.7966} & 0.8086          & 0.8577          & 0.8278          & 0.7998          & 0.7224          & 0.8211           \\
                    & Prob-UNet    & 0.8742          & 0.7389          & 0.8332          & 0.8495          & 0.8531          & 0.8537          & 0.7895          & 0.8386           \\ 
\cline{2-10}
                    & CRNP (Ours)  & 0.8962          & 0.7787          & 0.8605          & 0.8637          & \textbf{0.8748} & \textbf{0.8736} & 0.7969          & 0.8615           \\
                    & CRNP* (Ours) & \textbf{0.8963} & 0.7811          & \textbf{0.8742} & \textbf{0.8850} & 0.8688          & 0.8692          & \textbf{0.8329} & \textbf{0.8758}  \\
\hline
\end{tabular}
}\end{center}
\vspace{-3mm}
\end{table}

\begin{table}[h]
\caption{The performance comparison of CRNP and different challenge models on both CT and MR segmentation of MMWHS dataset. The best results for each column are in bold. $\uparrow$ sign indicates the higher value the better.}
\vspace{-5mm}
\label{tab:mmwhs-challenge}
\begin{center}
\scalebox{0.85}{
\begin{tabular}{l|cc|cc}
\hline
        & \multicolumn{2}{c|}{CT}                              & \multicolumn{2}{c}{MR}                    \\
Models  & Dice $\uparrow$                    & Jaccard $\uparrow$                 & Dice $\uparrow$               & Jaccard $\uparrow$            \\ \hline
GUT     & 0.9080 & 0.8320 & 0.8630 & 0.7620 \\
KTH     & 0.8940      & 0.8100      & 0.8550 & 0.7530 \\
CUHK1   & 0.8900      & 0.8050      & 0.7830 & 0.6530 \\
CUHK2   & 0.8860      & 0.7980      & 0.8100 & 0.6870 \\
UCF     & 0.8790      & 0.7920      & 0.8180 & 0.7010 \\
SIAT    & 0.8490      & 0.7420      & 0.6740 & 0.5320 \\
UT      & 0.8380      & 0.7420      & 0.8170 & 0.6950 \\
UB1 & 0.8870      & 0.7980      & 0.8690 & 0.7730 \\
UB2 & -                        & -                        & 0.8740 & 0.7780 \\
UOE & 0.8060      & 0.6970      & 0.8320 & 0.7200 \\ \hline
Ours    & \textbf{0.9193}                         & \textbf{0.8486}
                         & \textbf{0.8758}                    & \textbf{0.7814}
                   \\ \hline
\end{tabular}
}\end{center}
\vspace{-5mm}
\end{table}

As shown in Tab.~\ref{tab:mmwhs}, our proposed CRNP and its ensemble version have 7 out of the 8 best Dice results on both CT and MR. On CT (Tab.~\ref{tab:mmwhs}), CRNP raises LV Dice score from 0.9297 to 0.9369 and PA Dice score from 0.8425 to 0.8628, when compared to the second-best models. On whole heart segmentation Dice score, CRNP outperforms the second-best model by 1.9\%. The ensemble version of CRNP further improves segmentation accuracy.
A similar result is observed on MR. 
On LV, CRNP raises the Dice score from 0.8850 to 0.8962 and AA Dice score from 0.8551 to 0.8736 when compared to the second-best models. On whole heart segmentation, CRNP increases MR Dice from 0.8535 to 0.8615. Model ensemble further improves the performance. 

Interestingly, the Individual model obtains accurate results on CT (0.8989 for WH score). However, performance (0.6729 for WH score) drops drastically on MR evaluation, with particularly poor accuracy on RV, AA and PA. But when considering both modalities (3D UNet model), the model performance increases substantially. This shows the bounds of considering a single modality, especially for MR segmentation. The proposed CRNP outperforms the 3D Unet by a large margin. Ummkd \cite{dou2020unpaired}  performs consistently well on Myo on both CT and MR. We hypothesize that the domain-specific normalization and knowledge distillation loss contribute more to Myo segmentation than to other organs. Probability UNet tries to model posterior latent space rather than a deterministic prediction, which may explain its performance. In general, we note that the CT segmentation results are better than MR, which resonates with the conclusion from~\cite{zhuang2019evaluation}.

From the number of parameters  perspective, the randomized network is made up of 3 convolutional hidden layers and the prediction network has 2 convolutional hidden layers. So the change in number of parameters is minimal. More specifically, the number of parameters of competing methods are: 1) UNet: 41.05M, 2) Ummkd (with UNet backbone for fair comparison): 41.05M, and 3) ProbUNet: 57.44M. Our CRNP has 42.18M parameters, where the RNP module has 0.29M, and the attention module has 0.84M parameters.

We also compare the proposed CRNP model with the state-of-the-art models reported by the official challenge report~\cite{zhuang2019evaluation}. The results are shown in Tab.~\ref{tab:mmwhs-challenge}. On whole heart segmentation, CRNP has a particularly accurate Dice score and Jaccard index for CT and MR. Compared to the second-best models, our CRNP model increases the Dice score from 0.9080 to 0.9193 and from 0.8740 to 0.8758 on CT and MR, respectively. Similar results are shown forJaccard index.

\vspace{1mm}
\noindent\textbf{Performance on BraTS2020 dataset.} Developing automated segmentation models to delineate intrinsically heterogeneous brain tumors is the main goal of BraTS2020 Challenge. Following~\cite{wang2021transbts}, we compare the proposed CRNP model with many other strong methods, including 3D UNet \cite{cciccek20163d}, Basic VNet \cite{milletari2016v}, Deeper VNet \cite{milletari2016v}, Residual 3D UNet, Modal-Pairing \cite{wang2020modality}, TransBTS \cite{wang2021transbts}, as well as uncertainty-aware models ProbUNet \cite{kohl2018probabilistic} and SSN \cite{monteiro2020stochastic} that models aleatoric uncertainty by considering spatially coherence. We evaluate the Dice and Hausdorff95 indexes of all models on four organs: enhancing tumor (ET); tumor core (TC) that consists of ET, necrotic and nonenhancing tumor core; and whole tumor (WT) that contains TC and the peritumoral edema.

\begin{table}[t!]
\caption{The performance of different models on BraTS2020 Online validation set. The best results for each column are in bold. $*$ indicates models with ensemble. $\uparrow$ sign indicates the higher value the better; while $\downarrow$ means the lower value the better.}
\vspace{-4mm}
\label{tab:brats-performance}
\begin{center}
\scalebox{0.85}{
\begin{tabular}{l|ccc|ccc}
\hline
                                       & \multicolumn{3}{c|}{Dice $\uparrow$}                & \multicolumn{3}{c}{Hausdorff95 $\downarrow$}         \\
Models                                 & ET              & WT              & TC              & ET               & WT              & TC              \\ \hline
3D UNet \cite{cciccek20163d}           & 0.6876          & 0.8411          & 0.7906          & 50.9830          & 13.3660         & 13.6070         \\
Basic VNet \cite{milletari2016v}       & 0.6179          & 0.8463          & 0.7526          & 47.7020          & 20.4070         & 12.1750         \\
Deeper VNet \cite{milletari2016v}      & 0.6897          & 0.8611          & 0.7790          & 43.5180          & 14.4990         & 16.1530         \\
Residual 3D UNet                       & 0.7163          & 0.8246          & 0.7647          & 37.4220          & 12.3370         & 13.1050         \\
ProbUNet \cite{kohl2018probabilistic}  & 0.7392          & 0.8782          & 0.7955          & 36.2458          & 6.9518          & 7.7183          \\
SSN \cite{monteiro2020stochastic}      & 0.6795          & 0.8420          & 0.7866          & 43.6574          & 14.6945         & 19.5171         \\
Modal-Pairing* \cite{wang2020modality} & 0.7850          & 0.9070          & 0.8370          & 35.0100          & 4.7100          & 5.7000          \\
TransBTS \cite{wang2021transbts}       & 0.7873          & 0.9009          & 0.8173          & \textbf{17.9470} & 4.9640          & 9.7690          \\ \hline
CRNP (Ours)                            & 0.7887          & 0.9086          & 0.8372          & 26.5972          & \textbf{4.0490}          & 6.0040          \\
CRNP* (Ours)                           & \textbf{0.7902} & \textbf{0.9109} & \textbf{0.8550} & 26.4682          & 4.1096 & \textbf{5.3337} \\ \hline
\end{tabular}
}\end{center}
\vspace{-5mm}
\end{table}

In Tab.~\ref{tab:brats-performance}, our models have 5 out of the 6 best results. 
The CRNP improves the ET Dice score, compared with the second-best model, from 0.7873 to 0.7887; and from 0.9070 to 0.9086 on WT. Similar results are shown on Hausdorff95 indexes. 
Note that the Modal-Pairing model adopts an ensemble strategy. When applying the ensemble strategy to CRNP, the results improved even further. 
The WT Dice of CRNP* can reach 0.9109; the TC Dice can reach 0.8550, which is one more percent increment; and improves the TC Hausdorff95 to 5.3337. The performance improvements show the effectiveness of the proposed CRNP model.

\subsection{Computer Vision Classification Model Performance}

In this section, we show results that demonstrate the effectiveness of CRNP on multiple CV classification tasks. The evaluation metrics include accuracy and multi-class AUROC on Handwritten, CUB and Scene15 datasets. Following Han et al.~\cite{han2021trusted}, the comparison models include multiple uncertainty-aware models: Monte Carlo dropout (MCDO) \cite{gal2015bayesian} that adopts dropout at inference as a Bayesian approximator; deep ensemble (DE)~\cite{lakshminarayanan2017simple}, which uses an ensemble strategy to reduce uncertainty; uncertainty-aware attention (UA)~\cite{heo2018uncertainty} that creates uncertainty attention maps from a learned Gaussian distribution; evidential deep learning (EDL)~\cite{sensoy2018evidential} that predicts an extra Dirichlet distribution for all logits based on evidence; and trusted multi-view classification (TMC)~\cite{han2021trusted}, which is a multi-view version of EDL.

\begin{table}[t!]
\caption{The performance of different models on computer vision classification datasets. The best results for each row are in bold.}
\vspace{-4mm}
\label{tab:cv}
\begin{center}
\scalebox{0.85}{
\begin{tabular}{l|l|ccccc|c}
\hline
Data                         & Metric & MCDO \cite{gal2015bayesian}   & DE \cite{lakshminarayanan2017simple}     & UA \cite{heo2018uncertainty}     & EDL \cite{sensoy2018evidential}    & TMC \cite{han2021trusted}    & CRNP   \\ \hline
\multirow{2}{*}{Handwritten} & Acc    & 0.9737 & 0.9830 & 0.9745 & 0.9767 & 0.9851 & \textbf{0.9925} \\
                             & AUROC  & 0.9970 & 0.9979 & 0.9967 & 0.9983 & \textbf{0.9997} & 0.9996 \\ \hline
\multirow{2}{*}{CUB}         & Acc    & 0.8978 & 0.9019 & 0.8975 & 0.8950 & 0.9100 & \textbf{0.9167} \\
                             & AUROC  & 0.9929 & 0.9877 & 0.9869 & 0.9871 & 0.9906 & \textbf{0.9961} \\ \hline
\multirow{2}{*}{Scene15}     & Acc    & 0.5296 & 0.3912 & 0.4120 & 0.4641 & 0.6774  & \textbf{0.7057} \\
                             & AUROC  & 0.9290 & 0.7464 & 0.8526 & 0.9141 & 0.9594  & \textbf{0.9734} \\ \hline
\end{tabular}
}\end{center}
\vspace{-7mm}
\end{table}

As shown in Tab.~\ref{tab:cv}, CRNP model can outperform its counterparts on 5 out of 6 measures across datasets. CRNP performs particularly well on Scene15, increasing the accuracy from 0.6774 to 0.7057 (a 2.83\% improvement) and AUROC from 0.9594 to 0.9734 (a 1.4\% improvement). CRNP also has promising results on Handwritten and CUB data. 
On AUROC of Handwritten, CRNP gets slightly worse but comparable results than TMC (0.9996 vs. 0.9997).

\subsection{Ablation Study}
\subsubsection{Effectiveness of Each Component}
\label{sec:ablation-components}

In the ablation study, we examine each component of the proposed CRNP. The ``Base'' model is the plain multi-modal 3D UNet with dual branches; ``CA'' means cross-attention by assigning the query from one modality, while keep the key and value the other modality; ``SA'' means applying self-attention as we propose.
We conducted the ablation on the validation set split of the MMWHS dataset and we measured the average Dice scores of each organ on CT and MR. As shown in Tab.~\ref{tab:ablation-components}, compared with the Base 3D UNet model, the CRNP model is able to improve (around 1\% increment of Dice scores) the performance across multiple organs, where the improvements are especially obvious on Myo, LA, RA, AA and WH.  From the table, we can perceive that, with the help of either cross-attention or self-attention, the model performance can be further boosted. But applying the self-attention as described in Sec.~\ref{sec:cross-rnp}, causes the model to produce the best results (6 best results out of 8)  across multiple organs. This is mainly because the self-attention on the multi-modal feature fusion not only models the cross-modal relations, but also considers uni-modal attentions.

\begin{table}[t!]
\caption{Ablation study on MMWHS dataset. Best results per row are in bold.}
\vspace{-5mm}
\label{tab:ablation-components}
\begin{center}
\scalebox{0.85}{
\begin{tabular}{l|cccccccc}
\hline
Models  & LV     & Myo    & RV     & LA     & RA     & AA     & PA     & WH     \\ \hline
Base    & 0.9334 & 0.8596 & 0.8876 & 0.8932 & 0.8794 & 0.8239 & 0.8168 & 0.8706 \\
CRNP    & 0.9324 & 0.8685 & 0.8644 & 0.9007 & 0.8957 & \textbf{0.9216} & 0.8225 & 0.8865 \\
CRNP+CA & 0.9323 & 0.8683 & 0.8802 & 0.9147 & \textbf{0.9116} & 0.9098 & 0.8194 & 0.8909 \\
CRNP+SA & \textbf{0.9356} & \textbf{0.8891} & \textbf{0.8814} & \textbf{0.9232} & 0.8987 & 0.9148 & \textbf{0.8277} & \textbf{0.8958} \\ \hline
\end{tabular}
}\end{center}
\vspace{-2mm}
\end{table}

\vspace{-5mm}
\subsubsection{Discussion of Different CRNP Fusion functions}

In terms of different CRNP fusion functions that can be applied in $\operatorname{fusion}(.,.)$ (Sec.~\ref{sec:cross-rnp}), we compare and discuss three types, as shown in Tab.~\ref{tab:ablation-fusion}: (a) ``Replace'' represents a naive replacement of the original modality features by the uncertainty map 
attended features; (b)  ``Concat'' applies the concatenation operation on the original modality features and the uncertainty map 
attended features; and (c)  ``Residual'', which is the default fusion strategy of the proposed CRNP, denotes an addition operation performed between two feature tensors. This experiment is conducted on the MMWHS dataset and averages both CT and MR Dice results.
From the results, we note that all three types of fusion functions have pros and cons. However, the ``Residual'' model performs better (4 best results out of 8) than other functions. This advantage is more noticeable on RA, AA and PA, on which more than 1\% improvement is gained on Dice score.

\begin{table}[t!]
\caption{Analysis of different fusion functions of CRNP on MMWHS dataset. Best results per row are in bold.}
\vspace{-5mm}
\label{tab:ablation-fusion}
\begin{center}
\scalebox{0.85}{
\begin{tabular}{l|cccccccc}
\hline
Models   & LV     & Myo    & RV     & LA     & RA     & AA     & PA     & WH     \\ \hline
Replace  & \textbf{0.9342} & \textbf{0.8688} & 0.8688 & 0.897  & 0.8812 & 0.9074 & 0.8128 & 0.8815 \\
Concat   & 0.9327 & 0.8676 & \textbf{0.8798} & \textbf{0.9031} & 0.8781 & 0.9098 & 0.8042 & 0.8822 \\
Residual & 0.9324 & 0.8685 & 0.8644 & 0.9007 & \textbf{0.8957} & \textbf{0.9216} & \textbf{0.8225} & \textbf{0.8865} \\ \hline
\end{tabular}
}\end{center}
\vspace{-5mm}
\end{table}

\subsection{Visualization}

\begin{figure}[h]
\vspace{-2mm}
\begin{center}
\includegraphics[width=1.0\textwidth]{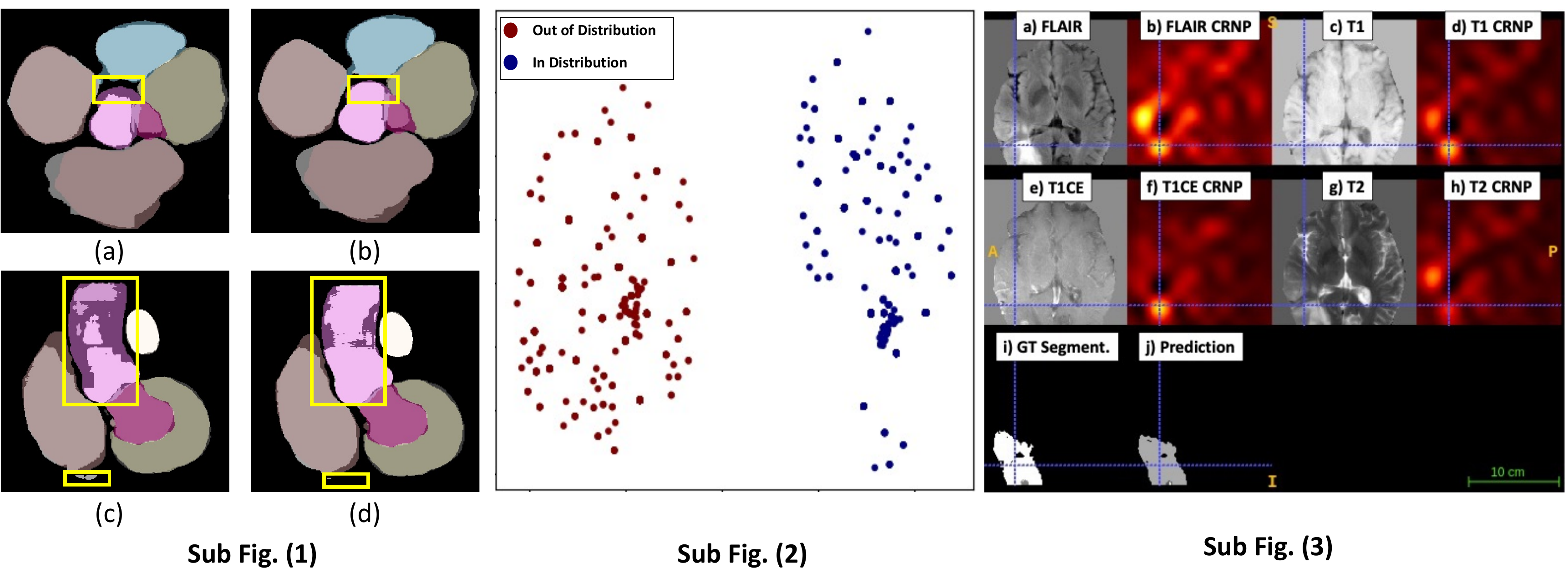}
\end{center}
\vspace{-6mm}
\caption{Visualization experiments of CRNP. Sub Fig.(1) shows a comparison between the segmentation of the proposed CRNP ((b) and (d)) and its Base model ((a) and (c)). 
Sub Fig.(2) shows the T-SNE graph of the in and out of distribution data points produced by the cross-modal RNP module. In the Sub Fig.(3), we show the CRNP uncertainty heat-maps.}
\label{fig:visualization}
\vspace{-3mm}
\end{figure}

We also conduct a visualization experiment in Fig.~\ref{fig:visualization} that shows the MMWHS segmentation visualization (Sub Fig.1), 
T-SNE visualization of in and out of distribution data points produced by the uncertainty maps from the RNP module on the CT images from MMWHS (Sub Fig.2), and the CRNP uncertainty heat-maps for BraTS2020 images (Sub Fig.3). 
As the two cases from validation set shown in Sub Fig.(1), (a) (c) are segmented by the Base model and (b) (d) are from CRNP. 
The color masks 
denote the 
segmentation results (e.g., pink) overlaid on the ground truth (e.g., purple). The obvious segmentation differences are highlighted by yellow boxes. When comparing segmentation from two models, we can notice that our CRNP has better segmentation results, especially on the organ edges. 
This is mainly because organ edges contain more uncertain regions. 
The proposed CRNP can perceive uncertain segmented regions within one modality and assign more weights to the other one.
By leveraging this information, CRNP is able to alleviate segmentation uncertainties in organ edges.
Moreover, we visualize the in and out of distribution uncertainty maps processed by T-SNE in Sub Fig.(2). Following Han et al.~\cite{han2021trusted}, we consider the original features as the in distribution data and noisy features modified by additive Gaussian noise as the out of distribution data. Then, these samples are fed into the cross-modal RNP modules to get the uncertainty map predictions. The T-SNE is able to clearly split these uncertainty predictions into two clusters. This shows further evidence of the effectiveness of our CRNP model to estimate uncertainties.
In Sub Fig.(3), we show the CRNP uncertainty heat-maps for a BraTS image, where the maps are estimated in the feature space and mapped back to the original image space. 
In this figure,  (a)(c)(e)(g) are the flair, t1, t1ce and t2  modalities; (b)(d)(f)(h) are the CRNP uncertainty maps for the modalities above (brighter pixel = higher uncertainty); and (i)(j) are the ground truth (GT) segmentation and CRNP prediction. 
Note that the high uncertainty regions are concentrated around the areas with brain tumors, which is reasonable since tumors are sparsely represented in the feature space, resulting in a large difference between RNP's random and prediction networks. 
Also note that the flair image has a stronger tumor signal than the other modalities, producing a larger uncertainty for the other modalities. In particular, this larger uncertainty will notify the other modalities to pay more attention to these areas.

\section{Conclusions}

In this paper, we proposed the Uncertainty-aware Multi-modal Learning model, named Cross-modal Random Network Prediction (CRNP). CRNP measures the total uncertainty in the feature space for each modality to better guide multi-modal fusion. Moreover, technically speaking, the proposed CRNP is the first approach to explore random network prediction to estimate uncertainty and fuse multi-modal data.
CRNP has a stable training process compared with a recent multi-modal approach that uses potentially unstable covariance measures to estimate uncertainty~\cite{monteiro2020stochastic}, and CRNP can also be easily translated between different prediction tasks.
Through experiments on two medical image segmentation datasets and three computer vision classification datasets, the effectiveness of the proposed CRNP model is verified. Also, ablation and visualization studies further validate CNRP as an effective multi-modal analysis method.



\clearpage
%
%
\bibliographystyle{splncs04}
\bibliography{mybib}
\end{document}